# A Cognitive Science perspective for learning how to design meaningful user experiences and human-centered technology

**Sara Kingsley**
Perspectives in Cognitive Science
Carnegie Mellon University
Pittsburgh, Pennsylvania
skingsle@cs.cmu.edu

## ABSTRACT
UPDATED—June 2, 2021. This paper reviews literature in cognitive science, human-computer interaction (HCI) and natural-language processing (NLP) to consider how analogical reasoning (AR) could help inform the design of communication and learning technologies, as well as online communities and digital platforms. First, analogical reasoning (AR) is defined, and use-cases of AR in the computing sciences are presented. The concept of schema is introduced, along with use-cases in computing. Finally, recommendations are offered for future work on using analogical reasoning and schema methods in the computing sciences.

## INTRODUCTION
Human-computer interaction (HCI) examines how people experience technology and technologically-mediated interactions, and how interactions affect user behavior, engagement and comprehension. Interacting with and retrieving information from online platforms constitutes a user experience. The challenge is that humans and machines perceive information in different ways [11]. Perception impacts inferences, behavior, and our understanding of the digital content that we consume.

Divergent interpretations about common events, objects or narratives invite opportunities for new ideas to emerge, as well as false notions [2, 5, 11, 15, 19]. Misinterpreted or misleading stories or facts are known to "go viral" and to increase the likelihood for incivility [11]. Referred to as "misinformation" or "disinformation," the phenomenon is, in part, a product of (exploiting) analogical reasoning and normal cognitive processes [3, 19]. Problematically, digital platforms are efficient mechanisms for spreading rumors, participating in misinterpretations, and for misconstruing fact-sharing as opinion [16]. Knowing how information manipulation and misinterpretation emerges from computer-mediated experiences, and as part of normal cognitive processes[9], is important for designing better user experiences and platforms.

Digital platforms develop algorithms to match content or users to users or groups of users (so-called "audiences"). The algorithms were designed to infer what content is relevant to whom. The decision requires algorithms to make additional layers of inferences or assumptions, including about:

- What is the user's profile? (demographics, socioeconomic status, behavioral traits)
- Which users are similar or dissimilar? (on what basis?)
- Which matches produce the maximum economic value (for whom)?
- Which users are optimal targets for high (low) price (quality) options/matches?
- What is a quality (for whom)?

A body of evidence suggests algorithms match irrelevant content to users, and this degrades the quality of the user experience for some on digital platforms. How could analogical reasoning inform and guide how we design algorithmically-managed systems?

Building on past research, this paper explores how "interpretation and perception" interact and produce misunderstanding or misinformation [11, 15, 16, 18]. Analogical reasoning is proposed as a mechanism to learn about incongruities between intended usability and how users actually experience technology.

## ANALOGICAL REASONING

**Analogical reasoning is a framework to think about how (mis)matches occur in digital environments.** In the cognitive science literature, analogical reasoning describes a process by which people map what is known about familiar objects, events, behaviors and lived beings and states (the source analog) to learn or make inferences about something unknown or unfamiliar (the target analog)[3, 6, 7, 8, 9, 18].

### Components and Forms of Analogical Reasoning
*Analogical Mapping*
Traditionally, analogical reasoning is modeled as: A : B :: C : D, or A is to B as C is to D[7, 9]. These analogies are called "proportional analogies" [9]. A, B and C represent objects or subjects and relations that have known or familiar components and meanings (the so-called source analog) [9]. D represents an unknown object or subject; the comparison between A and B allow us to infer the relationship between C and D. The inferred relationship between C and D (the target analog) allows us to learn about D [3, 9].

### Explaining Analogical Mapping via an Analogy
*Example 1: The Basketball Player*
Imagine there is a basketball player who never previously learned about or observed the game of soccer. The basketball player is given a soccer ball (the target). Without any prior information, the player might correctly infer the soccer ball is for playing a game and is used in a sport that has rules. However, the basketball player might wrongly assume the ball is a hand-sport (or they should use their hands to play with the ball). This inference is somewhat rational; basketballs (the source) are also round and the game of basketball requires players keep their hands on the ball while moving. Therefore, the basketball player's assumption is based on what they know about the game of and the object basketball. The player took what they know about a basketball and mapped that onto the soccer ball. This is an example of analogical reasoning. Analogical reasoning is imperfect in the sense that there is rarely a perfect match between components of each analog (i.e., how knowledge about a game and the shape of a known type of ball maps onto how one thinks about a different type of ball and how to use it).

In analogical reasoning, extraneous or irrelevant knowledge about a familiar object (the source, or basketball) is also mapped onto the unknown (the target, or soccer ball). What happens if the basketball player's mental model about the soccer ball is not updated or corrected? Would the basketball player continue using their hands to play with the ball? Would experience signal to the basketball player there is a better way to play ("user experience")? How could the player learn without any additional reference point to inform them about how to use the soccer ball? Is the weight or feel of the soccer ball enough to suggest using one's feet?

### Metaphor
Metaphor is a "abstract" or "symbolic" representation; these representations "are a special kind of analogy" [9]. Like analogies, metaphors leverage distinct domains to generate meaning, understanding or knowledge. The statement "I am getting slammed by final exams," is a metaphor that makes use of metonymy [9] or "substituting an associated concept" [a.k.a. "slamming a door"] into a different domain ("final exams") to convey something about that domain [9]. Metaphors are used to communicate about science, politics, medicine and to generally persuade and communicate about complex information [2, 9].

### Similarity
By definition, an analogy is a comparison of at least two patterns that have similar relations among their component parts. A foot (A) is to a human (B) what a paw (C) is to a dog (D). The analogy components (A, B, C, D) have a similar relations (part [foot, paw] is to a whole [human, dog]). A dis-analogy is when dissimilar relations or information components are mapped onto each other [9]. Disinformation campaigns on social media leverage imperfect comparisons or information patterns in a way that intentionally mislead users to perceive the pattern as coherent, legitimate, consistent or perfectly analogous to trustworthy patterns.

### Structure
The structure of analogy is important for knowledge representation and reasoning [9]. The "structure-mapping theory" of analogy states "analogy entails finding a structural alignment, or mapping, between domains" [9]. According to this theory, the two structures an analogy compares must have "consistent, one-to-one correspondences between mapped elements" [9]. Those familiar with linear algebra might compare the structural-completeness requirement to a inverse matrix where there is only one solution, or an image whose components map completely onto another image when transformed.

### Dis-analogies: Imperfect Comparisons
An alternative theory is analogies are inherently imperfect comparisons. Knowledge representation is rooted in and determined by context and environments [2, 3, 19]. In this way, knowledge representation is complex. Invisible components or domains are layered underneath surface or visible components of even seemingly simple representations. This paper argues that analyzing the dis-analogous or disjoint aspects of patterns in knowledge representations would enrich our understanding of how to design or re-design technology and user experience.

### Using Imperfect Information to Design User Experiences
User experiences on digital platforms may depend on analogical reasoning or faulty schema. Certainly, for example, online advertising platforms are less concerned about showing content that is valuable to every user than selling every user to valued content-providers [16]. Low quality digital ecosystems are created from such misalignment. Like the basketball player in the earlier example, platforms need mechanisms for learning from new information signals and for gathering additional data points to improve how different processes and operations function.

An author once described a scenario that speaks to this issue. The scene involved a classroom of young students who believed the earth was flat. In an experiment, the teacher informed the children that the earth is actually round. The children were then instructed to draw a picture of the earth. Most of the children drew the earth as a flat circle (versus a flat square or sphere). This experiment exampled how a teacher's mental model of how to convey information can mis-align with how the information is perceived by learners. The teacher may have produced the desired result by telling the students "the earth is not flat. It is a sphere like a basketball."

Unfortunately, in the aforementioned experiment, the teacher did not design a process to solicit the information to illuminate the fact that the students perceived the earth as "flat" and "square." The teacher needed to know this was how the students defined the earth (flat, square) prior to providing them with a contradicting or modifying model to instruct them about the shape of the world. If the students had been asked to draw the earth, first, the teacher might have learned the students perceived the planet as "flat" and "square." The knowledge would have enabled the teacher to think about how to modify the students' mental model of the earth from "flat" and "square" to "3-dimensional" and "circular." Perhaps, a short-cut would have been to say the earth is a "sphere like a basketball."

**Asymmetry: directional preferences in analogical reasoning**

Bowdle and Gentner (1997) argue that humans have strong preferences about where source and target domains are located in analogy [3]. In mathematics, this is like the order of operations of computing an equation: parentheses, exponents, multiplication and division (left to right), addition and subtraction (left to right). Bowdle and Gentner theorize that people prefer the (less-known/more-similar) target domain on the left side (A:B) of the statement (A:B::C:D), and the (more-known/less-similar) source domain on the right (C:D). Unfortunately, Bowdle and Gentner employ a racist example that compares North Korea to "Red China." The authors suggest that the research participants preferred the statement "North Korea is like Red China" versus "Red China is like North Korea" [3]. The authors theorized that this preference is explained by an assumption; namely that the participants believe North Korea is more similar to China than China is to North Korea. Of course, this inference seems culturally and politically-bound or rooted in the Cold War views of the United States. The United States political dogmas of the era might view China as "relatively communist." But this comparison collapses the complexity, diversity and richness of both places. How would people in China or North Korea measure or map the distance between similarity and difference?

**Analogical Transfer**
Analogical transfer is described as applying problem-solving methods from one domain and applying those methods to a new or different domain. The literature focuses on how to identify, construct and use analogies to empower people to: [i] solve problems outside their domain expertise; [ii] innovate; or, create new knowledge representations [10, 13, 14]. Despite this main focus on "positive", "consistent" or "perfect" pattern mapping, understanding how knowledge representations are mis-applied and mis-mapped onto domains is critical for effectively designing and understanding disinformation or propaganda campaigns, security and privacy in computing, natural language processing technologies, and learning experiences.

**Designing a system for Learning Meaningful Concepts**
L.H. Shu (2010) studied how to design a search engine that enables users to find meaningfully related concepts and works in biology [17]. For example, in a standard English dictionary and thesaurus, words listed as similar to or as synonyms for "clean" include: sanitary, hygienic, disinfected, dirt-free, decontaminated, aseptic, pure, unblemished and pristine. Words listed as dissimilar or as antonyms include: dirty, used, polluted, immoral, guilty, unfair, soiled, filthy, unclean. However, none of these word lists include "defend" as a word meaningfully related to "clean." L.H. Shu (2010) developed a method to discover meaningfully related words and concepts in biology by asking experts about keywords associated with words like "clean." The experts provided extra context that made apparent how "defend" was related to "clean" and for other word-pairs [17]. In biology, for example, we clean to defend against contamination or infection. In describing how the biomimemtric engine was designed, Shu illuminates that in specialized contexts or use-cases, languages take on other meanings, and component parts have different relationships. These extra or deeper meanings and relationships are important to capture when designing technology for specific communities.

*Analogies and Human-centered design*
User-studies about components or inputs to a system are critical for developing a comprehensive design process for technology [14, 17]. Shu (2010) shows how analogies are a method to collect data about the design of system inputs [17]. In Shu's case, standard English dictionaries proved inadequate for constructing keyword-pairs to base search engine results on. He discovered this by asking expert biologists for keywords that had meaningful associations to other words used in biology or their work [17]. Shu integrated the expert (user) feedback into the design of his system, fine-tuning his machine to produce search results that were meaningful for those working in or studying biological sciences.

In contrast to the human-centered biomimemtric search engine, contextual knowledge is intentionally collapsed in online advertising. Users are profiled and grouped together in "audiences." Predictions about user demographics, socioeconomic status and consumer habits feed into inferences and decisions that platforms (or algorithms) make about which users are similar or dissimilar, or classified as belonging to some identity or consumer group. The classifications (or the profiling) of users impact what content is shown and recommended to them. Platforms infer what users like or how users might behave based on past observations of the user, or based on other users who are identified as being similar.

The quantitative processes that produce these inferences don't always have "checks and balances" that allow for direct user feedback about whether the platform's assumptions align with the user's own identify, preferences, perceptions, narratives or experience of the world. For example, maybe a social media user lists their "home" as Massachusetts on their account homepage, but they actually live in Washington, DC. The user identifies Massachusetts as "home" because that is where they were born and raised and where their family lives. Washington, DC is where they have resided as an adult for work. So maybe the platform assumes showing the user employment opportunities in Massachusetts is most relevant. Since the user has turned off geo-location data for the app, there is no way for the platform to know where the user is really located. Beyond the irrelevance of the content shown to this user, the low-quality match wastes advertiser dollars by showing an employment ad to a consumer who will not apply for jobs in Massachusetts. What if the platform built a feature that asked consumers which locations they want to learn about employment opportunities? It could be as simple as allowing users to select which states are relevant to them for employment. This feedback system could enable the platform to create more accurate mappings between relevance and content and users. In addition, the un-selected options, in this case, could inform the platform about where inferences represent imperfect mappings (dis-analogies).

**which is better? Learning from accurate or inaccurate signals?**

**SCHEMA AND IMPROVING DIGITAL PLATFORMS**
Schema are another way to address and understand information problems in system feature, function and platform design [1]. A schema is a model, theory or plan that represents, outlines or organizes different system components and component relationships into a structure or system of categories and/or classifications [1, 8].[1]

Joseph Chee Chang et al. (2019) designed SearchLens, a tool that enables users to save, categorize and rank search result or information components ("factors") [12]. With SearchLens, users are able to "specify their interests and preferences" using keywords, and by ranking items that populate in a search [12]. Using "Lenses", SearchLens users decide which factors are most important [12]. For example, if browsing and consider what type of dog to adopt, with Lenses, the user could rank "doesn't shed a lot" as a very important factor compared to "dog size." Since users are able to define their own categories and relations between factors or information components, SearchLens allows us to learn about the mental models and preferences that users have, and how these models differ from that of the system designers' beliefs about users. Through its design, SearchLens intentionally solicits feedback about incongruities. As with the teacher who needed to learn their students' mental model of how the earth is shaped before trying to modify it, designers and developers also need to build mechanisms into systems that allow users to integrate how they would prefer to organize different factors or system components into their user experience. SearchLens examples how developers can do this: by building user feedback into the system as a feature rather than as an after-thought. Ultimately, by making user feedback a feature of the design process and system functionality, developers can learn what is needed for a technology to produce desired outcomes (for example, in the case of the school teacher, that users learn the earth is a 3D image in the shape of a circle or sphere).

---
[1]https://en.wikipedia.org/wiki/Schema(*psychology*)

**BUILDING INTELLIGENT MACHINES**
The task of building intelligent machines is often focused on "positive" pattern recognition. For example, signals or predictions that are correct versus wrong. In the computer sciences, a lot of effort is expended on learning from instances that were labeled by exogenous or thirty-party sources [11]. These labels train models how to make future predictions about new instances or events. The model parameters are tuned, but negative or mis-matching patterns are not always incorporated into the learning processes that intend to make machines intelligent. An exception might be in privacy and security in computing, where contradicting patterns inform model design. In positive or consistent pattern learning, the goal is to design machines that accurately or successfully complete human tasks by matching like-for-like. Among other things, these tasks include:

- Translation
- Transcription
- Tutoring or educating
- Predicting and Forecasting
- Classification and categorization
- Writing, Visual and Audio communication
- Providing customer service
- Aiding, helping or human care-taking

Negative or inconsistent pattern mapping helps identify gaps in systems or models that machines use. Inconsistent or sketchy patterns are useful for designing machines to complete tasks like:

- Vulnerability detection
- Identifying malicious agents and content [4, 5, 15, 16]
- Scraping un-indexed websites or finding missing data [4]
- Other forensics tasks

Learning from negative (or, inverted, sketchy or imperfect) patterns helps inform the design of positive patterns, and technologies that combat bias and discrimination. For example, audits of word-embeddings have unearthed invisible word associations that reflect bias or prejudices in society and impact model outcomes. Investigating and knowing about these negative patterns or components enables us to rethink the design of algorithmic systems and technology.

**CONTINUING AND FUTURE WORK**
Philology is the study of how languages morph over time in relation to changes in the political, cultural and socioeconomic ecologies in which languages are used. Building language-learning machines and systems allow us to learn about dis-analogies in user experiences – how mappings, meanings and relevance morph over time and between contexts [17, 11]. Disjoint semantic meanings and relationships morph depending on environmental, contextual and time-based factors. How meanings morph and misalign is meaning in of itself and a knowledge representation that could guide the development or re-design of algorithmic systems and technologies of the future [10]. Creating taxonomies and schema of mis-matches, mis-alignments, mis-perceptions and invisible data relations could illuminate how intelligent machines and systems should consider negative and positive patterns in inference and experience-making in a human-centered way.

**CONCLUSION**

Analogical reasoning and schema are useful mechanisms for learning about how users perceive information and the world around them. Research on analogical-thinking has focused on how to generate and leverage positive or perfect analogies in education, innovation and technology. Analogical reasoning, along with schema, are used to communicate and represent knowledge in politics, science, medicine and computing. However, analogies are imperfect representations of different knowledge domains. While at times, the imperfect nature of analogy is viewed as a limitation, investigating the imperfect nature of knowledge representations help illuminate open problems in natural-language processing, computer privacy and security, and in human-centered design.